\documentclass[lettersize,journal]{IEEEtran}
\usepackage{amsmath,amsfonts}
\usepackage{algorithmic}
\usepackage{algorithm}
\usepackage{array}
\usepackage[caption=false,font=normalsize,labelfont=sf,textfont=sf]{subfig}
\usepackage{textcomp}
\usepackage{stfloats}
\usepackage{url}
\usepackage{verbatim}
\usepackage{graphicx}
\usepackage{booktabs}
\usepackage{multirow}
\usepackage{cite}
\usepackage{amsthm}
\usepackage{amssymb}

\newtheorem{lemma}{Lemma}
\begin{document}

\title{Learning to Decode Linear Codes with Generalized Belief Propagation Network}

\author{
Samira Malek,
Pennsylvania State University, ~\IEEEmembership{Graduate Student Member,~IEEE}
\thanks {corresponding author: Samira Malek, Email: sxm6547@psu.edu, address: 201 Old Main, University Park, PA 16802.}
}


\maketitle
\begin{abstract}
Error correcting codes are a fundamental component of communication systems, enabling reliable data transmission over noisy channels. Recently, neural decoders have demonstrated notable performance gains over classical iterative algorithms. In this paper, we propose a Generalized Belief Propagation Network (GBPNet) that preserves symmetry conditions while enhancing its learning capability. The resulting architecture implicitly gives rise to fully connected transformations via the generalized variable-node updates across iterations, alongside learned element-wise scaling operation that enhance decoding performance, while retaining the Tanner graph structure. In addition, we introduce two effective training strategies—progressive training and high-SNR training—that significantly improve decoding accuracy. Experimental results on various linear codes show that the proposed GBPNet consistently outperforms conventional BP and state-of-the-art neural decoding.
\end{abstract}
\vspace{-3mm}
\section{Introduction}
\label{sec:intro}
{E}{rror} correcting codes play a fundamental role in reliable communication over noisy channels. Recently, deep learning-based decoders have shown promising performance, sometimes surpassing traditional decoding algorithms. Early approaches formulated decoding as a classification problem, where each codeword corresponds to a distinct class. However, because the number of codewords grows exponentially with the code length, the resulting classification task becomes computationally intractable for long codes. Consequently, such methods are generally limited to short codes \cite{wu2002neural,gruber2017deep}.

A syndrome-based deep learning framework was introduced in \cite{bennatan2018deep}, enabling training with a single reference codeword while maintaining identical performance across all codewords. Building on this idea, subsequent works employed more expressive architectures, including Transformers \cite{park2025crossmpt,choukroun2022error} and graph-attention networks \cite{qian2026learning}. However, these methods transform the received sequence into a syndrome-based representation and treat decoding primarily as an error-pattern prediction task, without explicitly exploiting the probabilistic dependencies and iterative message-passing structure of the Tanner graph.

Nachmani et al. \cite{nachmani2018deep} proposed a neural belief propagation (BP) decoder by unfolding BP iterations into a deep neural network. The method can be trained using only the all-zero codeword while explicitly exploiting the Tanner graph and iterative message-passing structure. Building on this framework, several BP-based neural decoders have been developed, including Hyper-Graph Network decoders \cite{nachmani2019hyper}, Tensor-Ring Neural BP decoders \cite{liang2024low}, and Adaptive Learned BP decoders \cite{tasdighi2025adaptive, 9163473}, which improve decoding performance through learnable message-update parameters.

Despite their success, existing BP-based neural decoders primarily rely on conventional BP. In contrast, GBPNet introduces generalized variable-node update rules that implicitly realize small fully connected blocks within the BP framework, together with learned scaling messages operation that enables richer message interactions across decoding iterations while preserving the Tanner graph structure. Furthermore, progressive training and high-SNR training strategies are developed to improve decoding performance. The symmetry condition is also proven to be preserved, enabling codeword-independent training. Experimental results on multiple linear codes demonstrate that GBPNet consistently outperforms conventional BP and state-of-the-art deep methods.

The remainder of this paper is organized as follows. After reviewing BP-based neural decoders in Section \ref{sec: background}, the generalized BP framework and the GBPNet architecture are presented in Section \ref{sec: GBP}. The symmetry condition of GBPNet is established in Section \ref{sec: SymmeteryCondition}, showing that training on the all-zero codeword yields identical performance for all codewords. Experimental results and comparisons with state-of-the-art methods are provided in Section \ref{sec: Exp}. Finally, conclusions are drawn in Section \ref{sec: conclusion}.

\section{Background}
\label{sec: background}
BP is an iterative message-passing algorithm defined on a Tanner graph, where messages are exchanged between variable and check nodes. Fig.~\ref{fig: TannerGraph} illustrates an example Tanner graph. Given the channel output $y_v$ and transmitted bit $x_v$, the channel loglikelihood ratio (LLR) is initialized as \cite{richardson2008modern}:
\begin{equation}
m_{v}^{ch} = \log \Big( \frac{\Pr(x_{v} = 1 \mid y_{v})}{\Pr(x_{v} = -1 \mid y_{v})}\Big ).
\end{equation}
A DNN form of BP was proposed in \cite{nachmani2018deep}, which preserves the BP message-passing structure while introducing trainable weights into the variable-node updates. The variable-to-check message at iteration $l$ is
\begin{equation}
   {m}_{v,c}^{(l)} = \operatorname{tanh}\Bigg(
\frac{1}{2}
\Big(
b_{v,c}^{(l)} m_{v}^{ch} +
\sum_{c' \in C_v \setminus \{c\}}
\omega_{v,c,c'}^{(l)} m_{c', v}^{(l-1)}
\Big)
\Bigg),
\end{equation}
where $C_v$ denotes the set of check nodes connected to variable node $v$, and $m_{c,v}^{(-1)}=0$ for $l=0$. The check-node update remains identical to conventional BP:
\begin{equation}
m_{c, v}^{(l)} =
2 \operatorname{arctanh}
\Bigg(
\prod_{v' \in V_c \setminus \{v\}}
m_{v', c}^{(l)}
\Bigg),
\end{equation}
where $V_c$ is the set of variable nodes connected to check node $c$. After $l$ iterations, the posterior LLR estimate is computed using Eq.~\eqref{eq: WBP-output}. The parameters $b_{v,c}^{(l)}$, $\omega_{v,c,c'}^{(l)}$, $b_v^{\text{out}}$, and $\omega_{c',v}^{\text{out}}$ are learned during training, while setting them all to one recovers the classical BP algorithm.
\begin{equation}
\label{eq: WBP-output}
o_v =
b_v^{\text{out}} m_v^{ch} +
\sum_{c' \in C_v}
\omega_{c',v}^{\text{out}} m_{c', v}^{(l)}.
\end{equation}
\section{Generalized Belief Propagation Network}
\label{sec: GBP}
We generalize the belief propagation network \cite{nachmani2018deep} by replacing the one variable-layer in each decoding iteration with $K$ cascaded variable sublayers, where $K\geq 1$. Let
\begin{equation}
v^{(0)},v^{(1)},\dots,v^{(K)}    
\end{equation}
denote the intermediate states across the variable-side transformation. For each edge $(v,c)$, define
\begin{equation}
m_{v^{(k-1)},v^{(k)},c}^{(l)}, \qquad k=1,\dots,K,    
\end{equation}
as the message produced by the $k$-th variable sublayer at iteration $l$. For $l=0$, we have:
\begin{equation}
m_{v,c}^{(0)}=m_v,
\end{equation}
\begin{equation}
m_{c,v}^{(0)}
=
2\operatorname{Arctanh}
\Bigg(
\prod_{v' \in V_c\backslash\{v\}}
\tanh\Big(\frac{m_{v',c}^{(0)}}{2}\Big)
\Bigg),
\end{equation}
For notational consistency, we set
\begin{equation}
m_{c,v,v^{(K)}}^{(0)} \triangleq m_{c,v}^{(0)}.
\end{equation}
For $l\geq 1$, the first variable sublayer is given by
\begin{equation}
\label{eq: vl-first}
\begin{aligned}
m_{v^{(0)},v^{(1)},c}^{(l)}
&=
\operatorname{tanh}\Bigg(
\frac{1}{2}\Big(
b_{v,c}^{(l,1)} m_v \\
&\quad +
\sum_{c'\in C_v\backslash\{c\}}
\omega_{v,c,c'}^{(l,1)}
m_{c',v,v^{(K)}}^{(l-1)}
\Big)
\Bigg)
\end{aligned}
\end{equation}
and for $k=2,\dots,K$,
\begin{equation}
\label{eq: vl-ks}
\begin{aligned}
m_{v^{(k-1)},v^{(k)},c}^{(l)}
&=
\tanh\Bigg(
\frac{1}{2}\Big(
b_{v,c}^{(l,k)} m_v \\
&\quad +
\sum_{c'\in C_v\backslash\{c\}}
\omega_{v,c,c'}^{(l,k)}
m_{v^{(k-2)},v^{(k-1)},c'}^{(l)}
\Big)
\Bigg)
\end{aligned}
\end{equation}
We further incorporate a learnable scaling layer after the last variable sublayer. 
Let $M^{(l)}$ denote the vector obtained by stacking all messages 
$m_{v^{(K-1)},v^{(K)},c}^{(l)}$ together with the channel messages 
$m_v^{\mathrm{ch}}$. Let $\mathcal{A}^{(l)}$ denote the element-wise scaling vector at 
iteration $l$.
\begin{equation}
\label{eq: vl-atten}
\tilde{M}^{(l)} = \mathcal{A}^{(l)} \odot M^{(l)},
\end{equation}
where $\odot$ denotes the Hadamard product. Denote by 
$\tilde{m}_{v^{(K-1)},v^{(K)},c}^{(l)}$ and $\tilde{m}_v^{\mathrm{ch}}$ 
the corresponding components of $\tilde{M}^{(l)}$. These reweighted messages 
are aggregated at the variable nodes as
\begin{equation}
\label{eq: vl-recommbination}
\begin{aligned}
\hat{m}_{v^{(K-1)},v^{(K)},c}^{(l)}
&=
\tanh\Bigg(
\frac{1}{2}\Big(
\tilde{m}_v^{\mathrm{ch}} \\
&\quad +
\sum_{c' \in C_v\setminus\{c\}}
\tilde{m}_{v^{(K-1)},v^{(K)},c'}^{(l)}
\Big)
\Bigg)
\end{aligned}
\end{equation}
The check-layer is then computed as
\begin{equation}
\label{eq: cl}
m_{c,v,v^{(K)}}^{(l)}
=
2\operatorname{Arctanh}
\Big(
\prod_{v' \in V_c\backslash\{v\}}
\hat{m}_{v'^{(K-1)},v'^{(K)},c}^{(l)}
\Big).
\end{equation}
\begin{figure}[!t]
\centering
\includegraphics[width=0.6\linewidth, height=0.091\textheight]{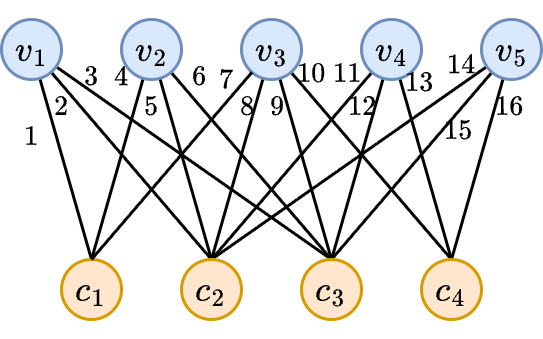}
\caption{Tanner graph with variable nodes $\{v_i\}_{i=1}^5$ and check nodes $\{c_i\}_{i=1}^4$.}
\label{fig: TannerGraph}
\end{figure}
\begin{figure}[!t]
\centering
\includegraphics[width=0.95\linewidth, height=0.3\textheight]{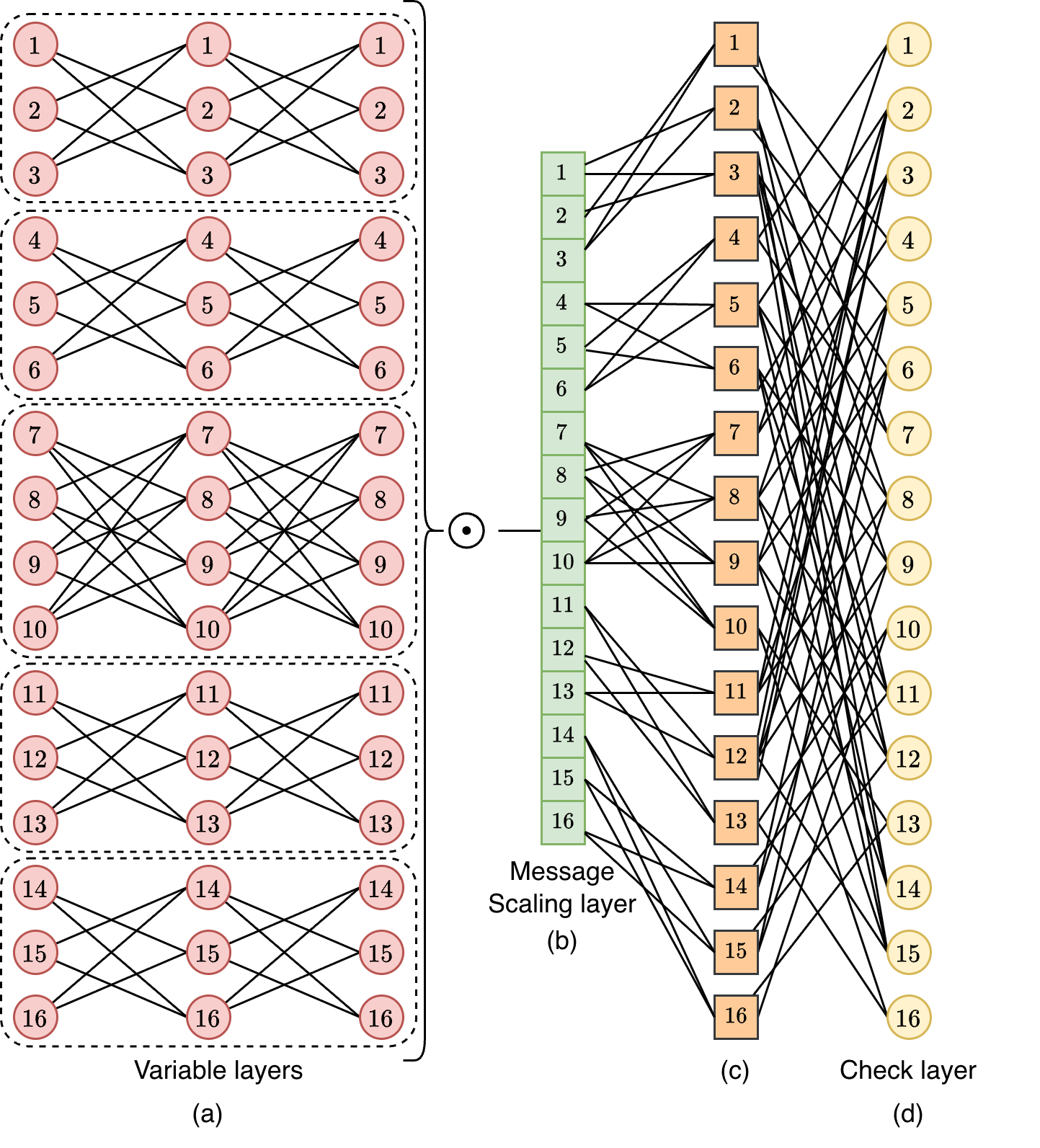}
\caption{Architecture of one GBPNet iteration for the Tanner graph in Fig.~\ref{fig: TannerGraph}, comprising three cascaded variable sublayers with five fully connected blocks, followed by a learned element-wise scaling operation layer, message recombination, and a check-node update.}
\label{fig: GBPNet}
\end{figure}
Fig.~\ref{fig: GBPNet} illustrates one iteration of GBPNet for the Tanner graph presented in Fig.~\ref{fig: TannerGraph}. The variable-node update is expanded into three cascaded sublayers composed of five small fully connected blocks, followed by a learned message-scaling layer and message recombination before the check-node update. The number of trainable parameters for the scaling layer is $E$ (the number of edges in the Tanner graph), and for each variable sublayer, it is $\sum d_{v_i}^2$, where $d_{v_i}$ is the degree of the $i$-th variable node. Therefore, each layer has $\sum d_{v_i}^2 \times K + E$ learnable parameters.
\section{Symmetery Condition}
\label{sec: SymmeteryCondition}
For block codes, standard message-passing symmetry conditions ensure codeword independent decoding error probability, allowing training on noisy realizations of a single codeword without loss of generality. This section establishes that the proposed generalized decoder preserves these conditions.

At iteration $l$, the variable-side computation consists of $K$ cascaded variable sublayers, followed by a scaling layer and a recombination layer. We represent this entire block by a single effective variable-node mapping
\begin{equation}
\label{eq: effective-variable-layer}
\Phi_v^{(l)}:
\left(m_v,\{m_{c',v,v^{(K)}}^{(l-1)}\}_{c'\in C_v\backslash\{c\}}\right)
\mapsto z_{v,c}^{(l)},
\end{equation}
where $z_{v,c}^{(l)}$ denotes the final message passed from variable node $v$ to check node $c$ at iteration $l$, while the effect of cascaded variable sublayers will be considered in Lemma \ref{lem: lemma1}. The check-node update is then given by
\begin{equation}
\label{eq: effective-check-layer}
m_{c,v,v^{(K)}}^{(l)}
=
2\operatorname{Arctanh}
\left(
\prod_{v' \in V_c\backslash\{v\}} z_{v',c}^{(l)}
\right).
\end{equation}
We next prove that $\Phi_v^{(l)}$ satisfies the variable-node symmetry condition and that the check-node update satisfies the check-node symmetry condition. As shown in \cite{nachmani2019hyper, richardson2008modern}, these two conditions are sufficient to ensure that the decoding error probability is independent of the transmitted codeword.
\begin{lemma}
\label{lem: lemma1}
The effective variable-node mapping $\Phi_v^{(l)}$ is odd and satisfies the variable-node symmetry condition:
\begin{equation}
\Phi_v^{(l)}\!\left(-m_v,\{-x_{c'}\}\right)
=
-\Phi_v^{(l)}\!\left(m_v,\{x_{c'}\}\right).
\end{equation}
\end{lemma}

\begin{proof}
Let
\[
f_1^{(l)}, f_2^{(l)}, \dots, f_K^{(l)}
\]
denote the $K$ cascaded variable sublayers, let $a^{(l)}$ denote the scaling block, and let $r^{(l)}$ denote the recombination layer. Then the overall variable-side mapping can be written as
\begin{equation}
\Phi_v^{(l)}
=
r^{(l)} \circ a^{(l)} \circ f_K^{(l)} \circ \cdots \circ f_2^{(l)} \circ f_1^{(l)}.
\end{equation}
For the variable sublayers in \eqref{eq: vl-first}, \eqref{eq: vl-ks}, and recommbination layer \eqref{eq: vl-recommbination}, Since there is no bias term, replacing $m_v$ and all incoming messages $m_{v^{(k-2)},v^{(k-1)},c'}^{(l)}$ by their negatives changes the argument of $\tanh$ by a factor of $-1$. Because $\tanh$ is odd, it follows that $f_i^{(l)}$ and $r^{(l)} $ are odd.

Next, the message scaling layer is given by
\begin{equation}
a^{(l)}(M^{(l)})=\mathcal{A}^{(l)}\odot M^{(l)}.
\end{equation}
Because $\mathcal{A}^{(l)}$ is a learned vector independent of $M^{(l)}$,
\begin{equation}
a^{(l)}(-M^{(l)})
=
\mathcal{A}^{(l)}\odot(-M^{(l)})
=
-\bigl(\mathcal{A}^{(l)}\odot M^{(l)}\bigr)
=
-a^{(l)}(M^{(l)}).
\end{equation}
Therefore, the message scaling block is odd.
Since the composition of odd mappings is odd, $\Phi_v^{(l)}$ is odd. Hence,
\[
\Phi_v^{(l)}\!\left(-m_v,\{-x_{c'}\}\right)
=
-\Phi_v^{(l)}\!\left(m_v,\{x_{c'}\}\right),
\]
which proves the variable-node symmetry condition.
\end{proof}
\begin{lemma}
The check-node update is given by \eqref{eq: effective-check-layer}. Then the check-node symmetry condition holds:
\begin{equation}
m_{c,v,v^{(K)}}^{(l)}(\{b_{v'}z_{v',c}^{(l)}\})
=
\left(\prod_{v' \in V_c\backslash\{v\}} b_{v'}\right)
m_{c,v,v^{(K)}}^{(l)}(\{z_{v',c}^{(l)}\}),
\end{equation}
for any $b_{v'}\in\{-1,+1\}$.
\end{lemma}
\begin{proof}
Let
\begin{equation}
 \Psi(z_1,\dots,z_r)
=
2\operatorname{Arctanh}\left(\prod_{i=1}^r z_i\right),
\qquad r=|V_c|-1.   
\end{equation}
Then for any sign pattern $b_1,\dots,b_r\in\{-1,+1\}$,
\begin{equation}
\begin{aligned}
\Psi(b_1 z_1,\dots,b_r z_r)
&= 2\,\operatorname{arctanh}\!\left(\prod_{i=1}^r b_i z_i\right) \\
&= 2\,\operatorname{arctanh}\!\left(
\left(\prod_{i=1}^r b_i\right)
\left(\prod_{i=1}^r z_i\right)
\right).
\end{aligned}
\end{equation}
Since $\operatorname{arctanh}(\cdot)$ is an odd function, Eq.~\eqref{eq: last-check-node-symetry} follows, satisfying the check-node symmetry condition.
\begin{equation}
 \Psi(b_1z_1,\dots,b_rz_r)
=
\left(\prod_{i=1}^r b_i\right)\Psi(z_1,\dots,z_r). 
\label{eq: last-check-node-symetry}
\end{equation}
\end{proof}
\section{Experiments}
\label{sec: Exp}
We evaluate GBPNet on BCH, Polar, LDPC, and MacKay codes, comparing its bit error rate (BER) performance with conventional BP and state-of-the-art neural decoders, as reported in Table~I. The parity-check matrix of the codes were obtained from \cite{channelcodes}. Training samples are generated by transmitting the all-zero codeword using BPSK modulation over an additive white Gaussian noise (AWGN) channel, justified by the message-passing symmetry of GBPNet. A total of $5\times10^7$ training samples are used, while testing employs random codewords until at least 500 frame errors are observed at each SNR. The network is optimized using Adam with a learning rate of $10^{-4}$ and a batch size of 200 using the cross-entropy multi-loss
in~\cite{nachmani2018deep}.

We further investigate two complementary training strategies: \textbf{high-SNR training} and \textbf{progressive training}. For high-SNR training, samples are generated over $4$--$10$ dB, while evaluation is performed over $1$--$8$ dB or $1$--$6$ dB, depending on the code. For progressive training, the network is initially trained with one decoding iteration. At each subsequent stage, a new iteration is randomly initialized, while previously trained iterations retain their learned weights. The entire network is then jointly trained for 200 epochs per stage until the target number of iterations is reached.

\begin{figure}[!t]
\centering
\includegraphics[width=0.97\linewidth]{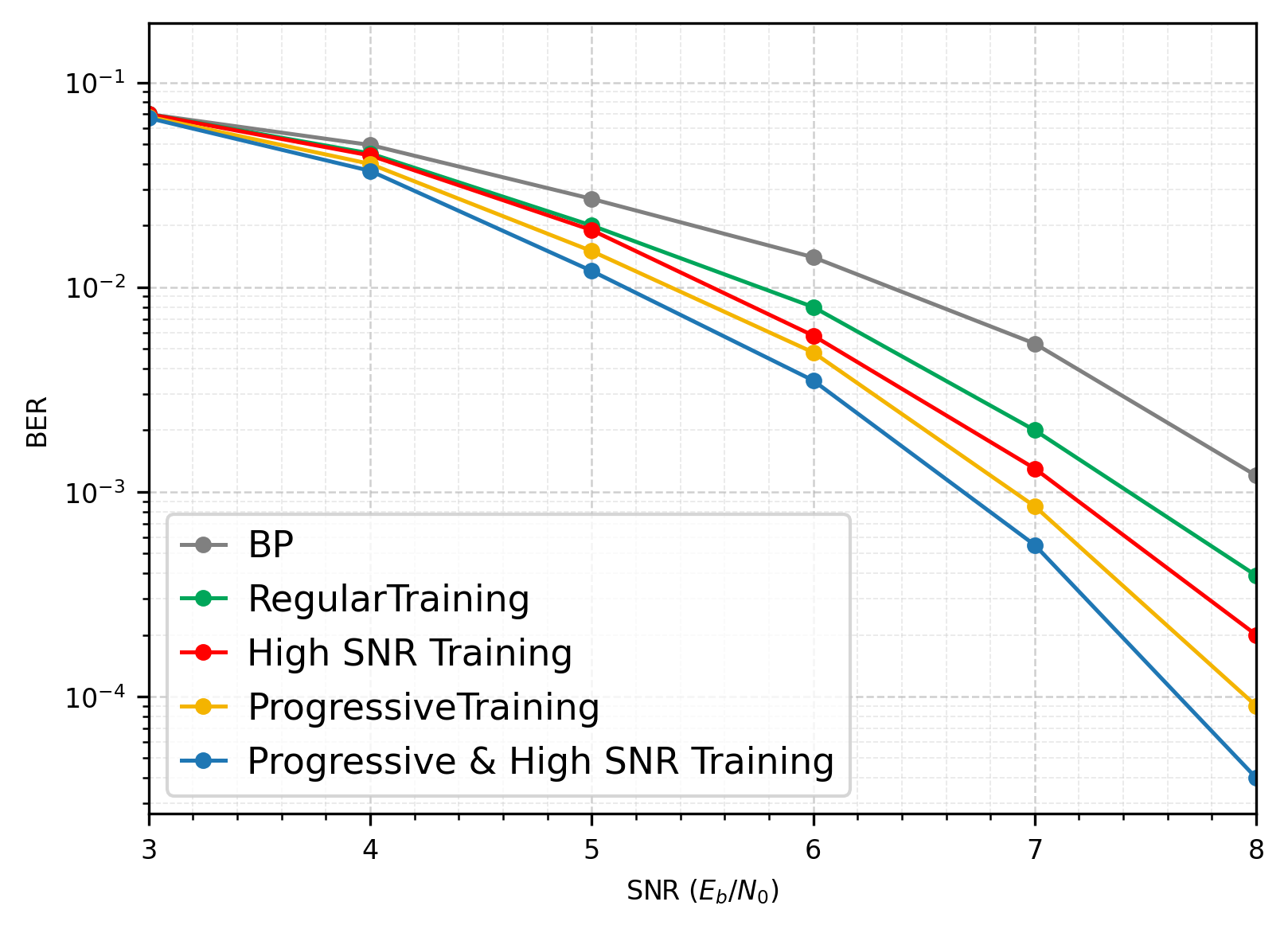}
\caption{Decoding performance comparison for BCH(127, 64), demonstrating that GBPNet outperforms conventional BP even with regular training, highlighting the effectiveness of its architecture, while high-SNR and progressive training provide further improvements.}
\label{fig: training-method}
\end{figure}

\begin{figure}[!t]
\centering
\includegraphics[width=0.97\linewidth]{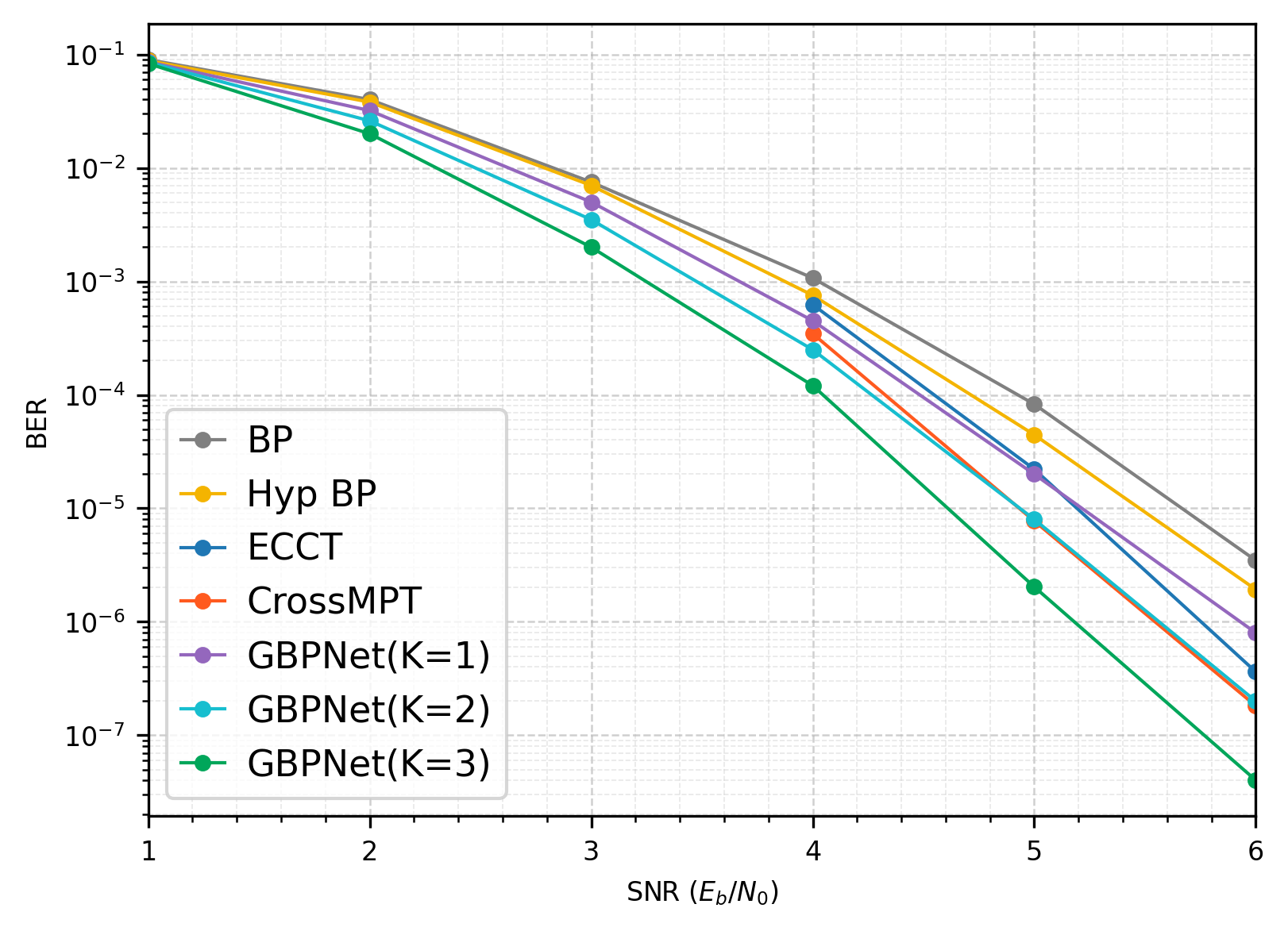}
\caption{BER comparison of different state-of-the-art decoding methods on the MacKay (96,48) code. The figure also shows the effect of using multiple variable-node sublayers, with GBPNet results for $K=1,2,3$.}
\label{fig: mackay}
\end{figure}

For all codes, the number of cascaded variable-node sublayers is fixed at $K=3$. GBPNet uses at most five decoding iterations, with $L=2$ for BCH codes, $L=3$ for Polar and LDPC codes, and $L=5$ for the MacKay $(96,48)$ code. For comparison, conventional BP and Hyp BP \cite{nachmani2019hyper} use five iterations, while the best reported results are used for the state-of-the-art transformer-based decoders ECCT \cite{choukroun2022error} and CrossMPT \cite{park2025crossmpt}. 
As shown in Fig.~\ref{fig: training-method}, GBPNet outperforms BP even with regular training, demonstrating the effectiveness of its architecture independently of the proposed training strategies. High-SNR and progressive training yield further BER improvements. Fig.~\ref{fig: mackay} additionally demonstrates GBPNet's competitive decoding performance against recent methods on the MacKay $(96,48)$ code. It also shows the effect of the number of cascaded variable-node sublayers, with GBPNet results for $K=1,2,3$, where increasing $K$ progressively improves the BER which is true for other codes as well.
\begin{table*}[t]
\label{tab: result}
\caption{Quantitative comparison of GBPNet at SNRs of $4$, $5$, and $6$ dB. Results are reported as $-\ln(\mathrm{BER})$ (larger is better).}
\centering
\small
\setlength{\tabcolsep}{4pt}
\begin{tabular}{l c ccc ccc ccc ccc ccc}
\toprule
\multirow{2}{*}{Codes} & \multirow{2}{*}{Parameter}
& \multicolumn{3}{c}{BP}
& \multicolumn{3}{c}{Hyp BP \cite{nachmani2019hyper}}
& \multicolumn{3}{c}{ECCT \cite{choukroun2022error}}
& \multicolumn{3}{c}{CrossMPT \cite{park2025crossmpt}}
& \multicolumn{3}{c}{GBPNet} \\
\cmidrule(lr){3-5}\cmidrule(lr){6-8}\cmidrule(lr){9-11}\cmidrule(lr){12-14}\cmidrule(lr){15-17}
& 
& 4 & 5 & 6
& 4 & 5 & 6
& 4 & 5 & 6
& 4 & 5 & 6
& 4 & 5 & 6 \\
\midrule

\multirow{2}{*}{BCH}

& (63,36) & 4.03 & 5.42 & 7.26 
& 4.29 & 5.91 & 8.01 
& 4.86 & 6.65 & 9.10 
& 5.03 & 6.91 & 9.37 
& \textbf{6.10} & \textbf{7.87} & \textbf{10.21} \\


& (63,51) & 4.50 & 5.82 & 7.42 
& 4.80 & 6.44 & 8.58 
& 5.66 & 7.89 & 11.01 
& 5.78 & 8.08 & 11.41 
& \textbf{6.42} & \textbf{8.78} & \textbf{12.03} \\

\midrule

\multirow{3}{*}{Polar}
& (64,32)  & 4.26 & 5.38 & 6.50 
& 4.59 & 6.10 & 7.69 
& 6.99 & 9.44 & 12.32 
& 7.50 & 9.97 & 13.31 
& \textbf{8.24} & \textbf{10.42} & \textbf{13.85} \\

& (64,48)  & 4.74 & 5.94 & 7.42 
& 4.92 & 6.44 & 8.39 
& 6.36 & 8.46 & 11.09 
& 6.51 & 8.70 & 11.31 
& \textbf{7.43} & \textbf{9.36} & \textbf{11.94} \\



& (128,96) & 4.61 & 5.79 & 7.08 
& 4.94 & 6.76 & 9.09 
& 6.31 & 9.12 & 12.47 
& 7.15 & 10.15 & 13.13 
& \textbf{7.96}& \textbf{10.97} & \textbf{13.89} \\

\midrule

\multirow{2}{*}{LDPC}

& (121,60) & 4.82 & 7.21 & 10.87 
& 5.22 & 8.29 & 13.00 
& 5.17 & 8.31 & 13.30 
& 5.74 & 9.26 & 14.78 
& \textbf{6.23} & \textbf{10.05} & \textbf{15.32} \\

& (121,70) & 5.88 & 8.76 & 13.04 
& 6.39 & 9.81 & 14.04 
& 6.40 & 10.21 & 16.11 
& 7.06 & 11.39 & 17.52 
& \textbf{8.12} & \textbf{12.21} & \textbf{18.32} \\


\midrule

MacKay
& (96,48) & 6.84 & 9.40 & 12.57 
& 7.19 & 10.02 & 13.16 
& 7.38 & 10.72 & 14.83 
& 7.97 & 11.77 & 15.52 
& \textbf{9.02} & \textbf{13.11} & \textbf{17.03} \\

\bottomrule
\end{tabular}
\end{table*}

\section{Conclusion}
\label{sec: conclusion}
We proposed GBPNet, a generalized neural BP decoder incorporating cascaded variable-node sublayers and a learned scaling messages block. We further introduced high-SNR and progressive training strategies. Experiments across various codes demonstrate improved BER performance over existing BP-based decoders. As future work, we aim to employ neural architecture search to optimize the number of variable-node sublayers in each iteration and the total number of decoding iterations.
\bibliographystyle{IEEEtran}
\bibliography{references}
\end{document}